\documentstyle[11pt,psfig,lingmacros,avm]{article}
\newcommand{\exm}{\em}
\newcommand{\aF}{\sf }

\newcommand{\aS}{\em }
\avmfont{\sf}
\avmvalfont{\it}
\avmsortfont{\small\it}

\begin{document}
\begin{center}
\Large\sc \ \\[2ex] Morphological Productivity\\ in the Lexicon\\[2ex]
\normalsize\rm
Onur T. \c{S}ehito\u{g}lu and H. Cem Bozsahin\\
Laboratory for the Computational Studies of Language\\
Department of Computer Engineering\\
Middle East Technical University, Ankara, Turkey\\
\tt \{onur,bozsahin\}@LcsL.metu.edu.tr\\[3ex]
\end{center}
\begin{abstract}

In this paper we outline a lexical organization for Turkish that makes
use of lexical rules for inflections, derivations, and lexical category
changes to control the proliferation of lexical entries. Lexical
rules handle changes in grammatical roles, enforce type constraints,
and control the mapping of subcategorization frames in valency changing
operations. A lexical inheritance hierarchy facilitates the enforcement
of type constraints. Semantic compositions in inflections and
derivations are constrained by the properties of the terms and predicates.

The design has been tested as part of a HPSG grammar for Turkish. In terms
of performance, run-time execution of the rules seems to be a far better
alternative than pre-compilation. The latter causes exponential growth
in the lexicon due to intensive use of inflections and derivations
in Turkish.
\end{abstract}

\section{Introduction}
\label{sect:intro}
Languages like Finnish, Hungarian, and Turkish have relatively rich
morphology which governs grammatical functions often delegated
to syntax in languages such as English. 
Prominence of morphology puts a greater demand on the information 
in the lexicon,
which may grow to an unmanageable size due
to heavy use of inflections and derivations. 
In Turkish, for instance, the nominal paradigm has
three affixes (number, case, relativizer), and the verbal paradigm has
eight (for voice, tense, person, aspect, and mood). Generating the full
paradigm for a nominal and a verbal root requires $2^3$ and $2^8$ entries
in the lexicon, respectively. The problem is further complicated by the rich
inventory of derivational affixes for both paradigms, as 
exemplified in~\ref{ex:intro1}. Hankamer~\cite{hankamer89} argues 
convincingly that full listing of every word form in the lexicon is 
untenable for agglutinative languages.

\enumsentence{\label{ex:intro1}\small
\shortex{3}{\exm Yaz-{\i}c{\i}-lar-a &\exm g\"{o}r-ev-ler-i &\exm bil-dir-il-me-mi\c{s}-ti}
 {write-\scriptsize VtoN-PLU-DAT & able-\scriptsize VtoN-PLU-ACC & 
know-\scriptsize CAUS-PASS-NEG-ASP-TENSE}
{'The clerks have not been informed of their duties'}
}

Handling inflections and derivations with lexical rules
opens us possibilities for encoding semantic and grammatical
changes in the lexicon as well. For instance, a causative
suffix will demote an agent to a patient or a recipient, and it will
add a new grammatical role for the causer (the new agent). 
A locative case suffix will mark a NP as an adjunct, which can no longer
satisfy subcategorization requirements of the verbs or postpositions.
We elaborate on the consequences of these phenomena in 
section~\ref{sect:types}. 

Another source for economy of representation can be seen in 
example~(\ref{ex:adj}), where attributive adjectives are used 
as nouns in \ref{ex:adj}b and \ref{ex:adj}d. One solution to this
problem is syntactic underspecification, e.g., grouping the nouns 
and adjectives under a single lexical category.\footnote{In fact,
traditional Turkish grammar books such as~\cite{lewis67} collectively
call them ``substantives.''} An alternative is to use
a lexical rule for differentiating predicate and term reading
of the lexical entry. 

\eenumsentence{
\label{ex:adj}
\item[a.]\shortex{2}{\exm kuru & \exm yaprak}
                    { dry & leaf}
                    {'dry leaf'}
\item[b.]\shortex{2}{\exm meyve & \exm kuru-su}
                    {fruit & dry-POSS}
                    {'dried fruit'}
\item[c.]\shortex{2}{\exm ya\c{s}-l{\i} &\exm  han{\i}m}
                    {age-ADJ & lady}
                    {'old lady'}
\item[d.]\shortex{2}{\exm b\"{u}t\"{u}n & \exm ya\c{s}-l{\i}-lar}
                    { all & age-ADJ-PLU}
                    {'all elderly'}
}

In what follows, we will describe different kinds of lexical rules
for type constraints, and 
handling changes in grammatical roles or subcategorization requirements.
We also discuss processing issues such as
run-time generation versus pre-compiling of word forms. 

\section{Morphology-syntax Interface}
Modelling inflections, derivations, and the corresponding phonological
alternations via lexical rules amounts to the lexicalization of morphology.
The alternatives to this approach (for Turkish) have also been explored,
e.g., the modularization of syntax and morphology by keeping them 
(and their lexicons) as
separate systems that communicate with each other~\cite{GunKO95},
or integrating morphology, syntax and semantics, thus treating 
morphotactics in the same manner as syntax with respect to
semantic composition~\cite{BozGoc95}. From a computational point of
view, the modular approach has efficient lexical access since lexical
search is performed on root forms, and bound morphemes are not considered
lexical items. In the integrated (multi-dimensional) approach, the lexicon
contains free and bound morphemes; they have complete syntactic and 
semantic specifications. Some of the inflections, e.g. person and number,
do not have any contribution to semantics, hence their semantic form (or LF) 
is that of identity. Some inflections, such as case and causative
affixes, compose semantic form of the stem (LF$_s$) with that
of the affix. LF$_s$ can be turned into ({\em cause} $x$ LF$_s$) for causatives
where $x$ is the new argument introduced by the causative affix.\footnote{cf.
example~\ref{ex:causative}} 
Similar arguments can be made for the semantic contribution of adjunct case
markers. 

The lexical approach to morphology presented here is a mid-point in the
design of the morphology-syntax interface. In this view, morphology is
not isolated from syntax, but, similar to the modular organization,
bound morphemes are not considered lexical items. They
can be attached to stems via lexical rules. This implies that lexical
rules are responsible for semantic composition and for the changes in
syntactic requirements. This view also represents a middle ground in
the complexity of lexical structures.

Keeping morphology and syntax entirely separate forces one to stipulate
different scopes for affixes. For instance, the adverbial suffix
{\exm -ken} and the adjectival {\exm -lu} might have 
phrasal (\ref{ex:scope}a and \ref{ex:scope}c) or lexical scope 
(\ref{ex:scope}b and \ref{ex:scope}d). Multi-dimensional approach
allows affixes to 'pick out'  different scopes in mixed 
morphological and syntactic composition. The lexical approach 
can accomodate both readings, provided that lexical rules are invoked 
with relevant syntactic information, e.g., valency of the verb. 
Morphologically ambiguous cases such as~\ref{ex:amb} are handled by
multiple instantiations of the lexical rules. 

\eenumsentence{\label{ex:scope}
\item[a.]\shortex{5}{\exm\c{C}ocuk &\exm top-a &[\exm kaleci-ye 
  &\exm bakar\mbox{\rm ]}-ken &\exm vurdu}
                    {child & ball-DAT & goalkeeper-DAT & look-ADV & hit}
                    {'The child hit the ball facing the goalkeeper.'}
\item[b.]\shortex{4}{\exm\c{C}ocuklar 
  &\exm \mbox{\rm [}y\"ur\"ur\mbox{\rm ]}-ken 
   &\exm ta\c{s} &\exm toplam\i\c{s}lar}
                    {children & walk-ADV & stone & picked}
                    {'The children had picked stones while walking.'}
\item[c.]\shortex{3}{\exm \mbox{\rm [}Uzun & \exm kol\mbox{\rm ]}-lu 
 &\exm g\"omlek}
                    {long & sleeve-ADJ & shirt}
                    {'shirt with long sleeves'}
\item[d.]\shortex{3}{\exm Uzun & \exm \mbox{\rm [}\c{c}i\c{c}ek\mbox{\rm ]}-li 
  & \exm g\"omlek}
                    {long & flower-ADJ & shirt}
                    {'long shirt with flower patterns'}
}

\enumsentence{\label{ex:amb}
\begin{tabular}[t]{llllll}
a. & \exm kalem-ler-i &b.&  \exm kalem-ler-i & c.&  \exm kalem-leri\\
   &      pencil-PLU-ACC & &     pencil-PLU-POSS.3SG & & pencil-POSS.3PL\\
   & 'the pencils (=OBJ)'& &     'his/her pencils' & & 'their pencils'\\
\end{tabular}
}

It is too early to evaluate
the advantages and disadvantages of these approaches in terms of
competence grammars and performance issues. But the choice of the strategy
also affects the design of lexical organization. For instance,
if inflections and derivations are handled by lexical rules, 
the morphological features need not be kept in the lexicon, since
the lexical rules will reflect the changes in syntactic and semantic
requirements coming from morphology. If morphology is treated almost 
like syntax, lexical
knowledge should contain richer morphological information, including
a semantic representation for bound forms (affixes), information
about boundedness/freeness of morphemes, and the type of attachment 
(e.g., affixation, cliticization, syntactic 
concatenation)~\cite{BozGoc95,hoeksemajanda88}. This will enable
the system to rule out, for instance, affixation of two free forms, 
or impose selectional restrictions on the stems of affixes.

\begin{figure}[hbt]
\centerline{\psfig{file=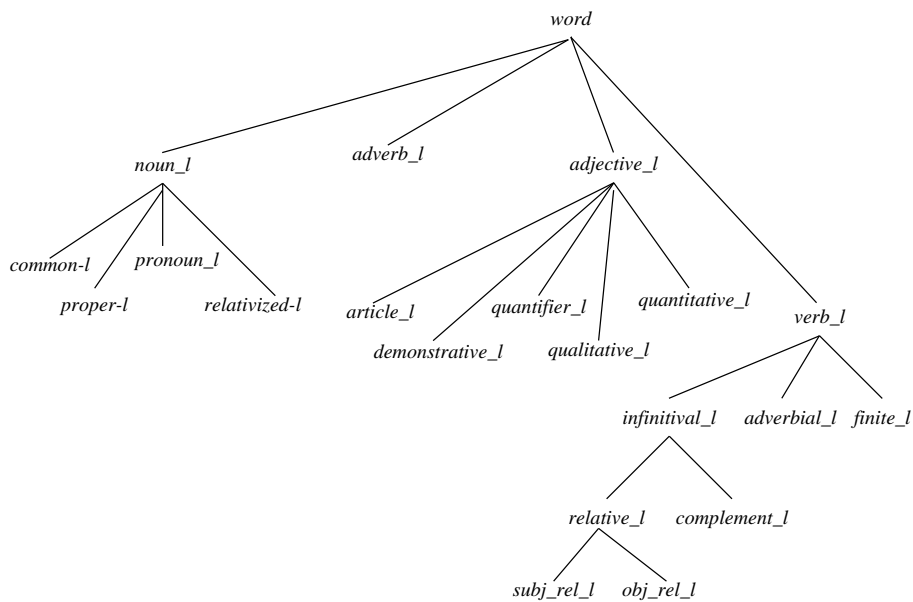}}
\caption{Lexical hierarchy}
\label{fig:wordhier}
\end{figure}

In this study, a lexical inheritance hierarchy is used in conjunction
with the lexical rules to obtain type constraints and feature 
structures for free forms (words); bound forms are not part of the lexicon.
The hierarchy is given in Figure~\ref{fig:wordhier}. 

This tree is part of a greater hierarchy which includes inheritance 
information for words and phrases. We make use of the inheritance and
type-checking mechanism of ALE~\cite{ALE} to impose type-specific
constraints on words. Words are distinguished from phrases by 
disallowing any kind of gapping below the word level in the tree. 
Designating a lexical item as one of the subtypes in the hierarchy
will apply all the constraints and incorporate the feature structures
of the supertypes along the path to {\em word}. For instance, a qualitative
adjective (e.g., {\exm rahat}=comfortable) is distinguished from a 
quantitative one (e.g., {\exm \c{c}ift}=double) by its choice of modifiers;
the latter does not allow intensifiers~(\ref{ex:qualadj}).

\eenumsentence{
\label{ex:qualadj}
\item[a.]\shortex{3}{\exm \c{c}ok & \exm rahat & \exm koltuk}
                    { very & comfortable & couch}
                    {'very comfortable couch'}
\item[b.]{\exm * \c{c}ok  \c{c}ift koltuk}
\item[c.]\shortex{3}{\exm rahat & \exm \c{c}ift & \exm koltuk}
                    { comfortable & double & couch}
                    {'comfortable twin couch'}
}

The fragments\footnote{We use HPSG style feature structures and signatures
in our descriptions. See Pollard and Sag~\cite{PolSag94}.}
of the type constraints for these subtypes are given in Figure~\ref{fig:cons}.
The controlled use of type constraints at different levels of the
lexical hierarchy eliminate the need to enumerate type-specific lexical rules
to achieve the same effect. 

\newsavebox{\tmp}
\newsavebox{\tmptwo}
\savebox{\tmp}{
{\scriptsize
\begin{avm}
 \[ 
			{\aF MODSYN} \; \[ {\aF LOCAL\|CAT} \; \[
			      {\aF HEAD} \; {\aS common} \\
			      {\aF ADJUNCTS} \; \[ {\aF QUANT} \; $-$ \\
						  {\aF QUANT-ADJ} \; $-$ \\
						  {\aF QUAL-ADJ} \; \@2 \\
						  {\aF NON-REF} \; \@3
					       \] 
							 \]
					\] \\
			{\aF MODADJ} \; \[  {\aF QUANT} \; $-$ \\
					    {\aF QUANT-ADJ} \; $+$ \\
					    {\aF QUAL-ADJ} \; \@2 \\
					    {\aF NON-REF} \; \@3
					\]
				      			\]
\end{avm}
}
}

\begin{figure}[hbt]
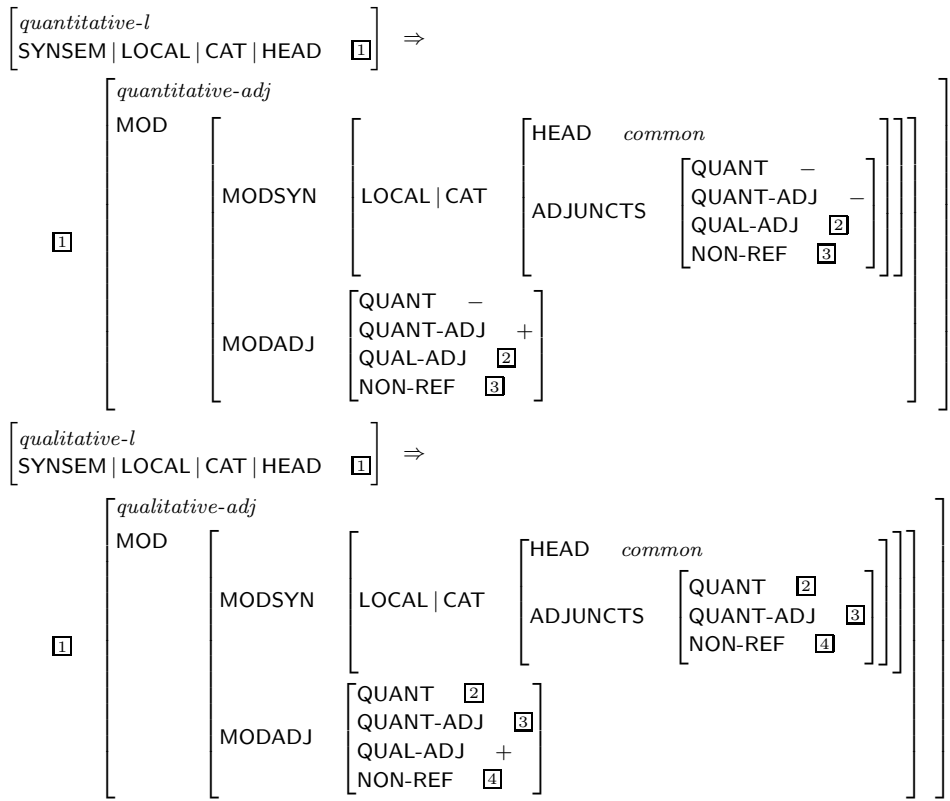

{\scriptsize
\begin{avm}
\avml
\[ {\aS quantitative-l} \\
   {\aF SYNSEM\|LOCAL\|CAT\|HEAD} \; \@1 \] \; $\Rightarrow$\\
   \hspace{0.5cm}
   \@1 \; \[ {\aS quantitative-adj} \\
             {\aF MOD} \; \usebox{\tmp}
	  \]
\avmr
\end{avm}
\savebox{\tmp}{
{\scriptsize
\begin{avm}
 \[ 
			{\aF MODSYN} \; \[ {\aF LOCAL\|CAT} \; \[
			      {\aF HEAD} \; {\aS common} \\
			      {\aF ADJUNCTS} \; \[ {\aF QUANT} \; \@2 \\
						  {\aF QUANT-ADJ} \; \@3 \\
						  {\aF NON-REF} \; \@4
					       \] 
							 \]
					\] \\
			{\aF MODADJ} \; \[  {\aF QUANT} \; \@2 \\
					    {\aF QUANT-ADJ} \; \@3 \\
					    {\aF QUAL-ADJ} \; $+$ \\
					    {\aF NON-REF} \; \@4
					\]
\]
\end{avm}
}
}
\begin{avm}
\avml
\[ {\aS qualitative-l} \\
   {\aF SYNSEM\|LOCAL\|CAT\|HEAD} \; \@1 \] \; $\Rightarrow$\\
   \hspace{0.5cm}
   \@1 \; \[ {\aS qualitative-adj} \\
             {\aF MOD} \; \usebox{\tmp}
	  \]
\avmr
\end{avm}
}

%
\caption{Type constraints for words and some subtypes.}
\label{fig:cons}
\end{figure}

\section{Types of lexical rules}\label{sect:types}
\paragraph{Inflections:} Lexical rules for inflections can check
morphotactic constraints for proper ordering of morphemes. 
More importantly, they should reflect the grammatical or semantic requirements
imposed by inflections. For instance, the locative case suffix in 
Turkish also marks an NP as adjunct~(\ref{ex:adjunct}).

\enumsentence{\label{ex:adjunct}
\shortex{3}{\exm Adam & \exm araba-da & \exm uyu-du}
           { man & car-LOC & sleep-TENSE.3SG}
           {'The man slept in the car'}.
}

\begin{figure}[hbt]
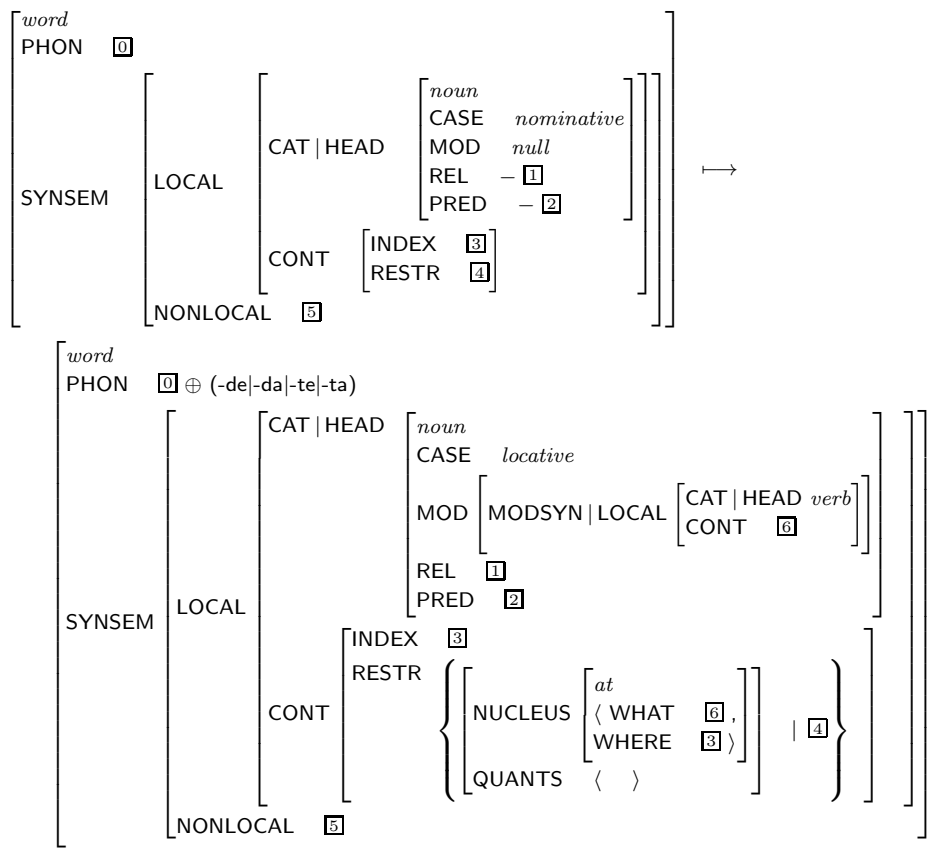


\savebox{\tmp}{
{\scriptsize
\begin{avm}
\[ {\aS noun} \\
   {\aF CASE} \; {\aS locative} \\
   {\aF MOD} \[ {\aF MODSYN\|LOCAL} \[ {\aF CAT\|HEAD} {\aS verb} \\
					     {\aF CONT} \; \@6
					  \]
		\] \\
   {\aF REL} \; \@1 \\
   {\aF PRED} \; \@2
\]
\end{avm}
}
}

\savebox{\tmptwo}{
{\scriptsize
\begin{avm}
\{ \[ {\aF NUCLEUS} \[ {\aS at} \\
			  \q< {\aF WHAT} \; \@6 ,\\
			  {\aF WHERE} \; \@3 \q>
		       \]\\
      {\aF QUANTS} \; \q< \; \q> 
    \] \; \| \@4 
\}
\end{avm}
}
}

{\scriptsize
\begin{avm}
\avml
\[ {\aS word} \\
   {\aF PHON} \; \@0\\
   {\aF SYNSEM} \; \[ {\aF LOCAL} \; \[ {\aF CAT\|HEAD} \;
				\[ {\aS noun} \\
				   {\aF CASE} \; {\aS nominative} \\
				   {\aF MOD} \; {\aS null} \\
				   {\aF REL} \; $-$ \@1 \\
				   {\aF PRED} \; $-$ \@2
				\]\\
					{\aF CONT} \; 
				\[ {\aF INDEX} \; \@3 \\
				   {\aF RESTR} \; \@4
				\]
				     \]\\
		      {\aF NONLOCAL} \; \@5
		   \]
\] \; $\longmapsto$\\
\hspace{0.5cm}
\[ {\aS word} \\
   {\aF PHON} \; \@0 $\oplus$ (-de$\mid$-da$\mid$-te$\mid$-ta)\\
   {\aF SYNSEM} \[ {\aF LOCAL} \[ {\aF CAT\|HEAD} \usebox{\tmp} \\
					{\aF CONT} 
				\[ {\aF INDEX} \; \@3 \\
				   {\aF RESTR} \usebox{\tmptwo}
				\]
				     \]\\
		      {\aF NONLOCAL} \; \@5
		   \]
\]
\avmr
\end{avm}
}
\caption{Lexical rule for the locative case.}
\label{fig:loc}
\end{figure}

The lexical rule for locative case is given (in ALE notation) in
Figure~\ref{fig:loc}. 
This rule is applied when the locative suffix is attached to a nominal stem.
The head of the NP is marked with the locative case,
and the type of NP is changed to an adjunct. 
This is achieved by modifying the head feature {\aF MOD}: While the nominative
marked noun has null value, a {\aF MODSYN} value with verbal head is
introduced in the head feature of the locative noun. This will allow
the locative marked noun to modify a verb.
Thus, it cannot satisfy
the subcategorization requirements of verbs or postpositions. This issue
is critical for parsing relatively free word-order languages where
grammatical relations are often indicated by overt case marking rather
than structural position. Figure~\ref{fig:loc} also shows the derivation
of the semantic representation for the case marked NP; {\em at(x,y)} is a
second-order predicate that holds between a term $x$ and a predicate $y$.
This predicate is inserted into the set of restrictions for the noun.
Although this method is not generative in the sense of ~\cite{puste91}, it
allows semantic composition in the lexicon.

\paragraph{Derivations:}
Denominal verbs, deverbal nouns, and part of speech changes can be modelled
respectively by adding subcategorization frames, discharging
subcategorization frames, and type coercions, via lexical rules.
The most difficult issue in derivations is the semantic composition.
For instance, the -CI morpheme (with allomorphs {\exm 
-c{\i}/-ci/-cu/-c\"{u}/-\c{c}{\i}/-\c{c}i/-\c{c}u/-\c{c}\"{u}}) adds the meaning
``doer/user of something''~(\ref{ex:cu}a), ``seller/lover of something''
~(\ref{ex:cu}b), or habitual~(\ref{ex:cu}c). 

\eenumsentence{\label{ex:cu}
\item[a.]\shortex{2}{\exm yol &\exm  -cu}
                    {road & }
                    {'traveller'}
\item[b.]\shortex{2}{\exm \c{s}eker & \exm -ci}
                    {candy & }
                    {'candy seller or lover'}
\item[c.]\shortex{2}{\exm sabah & \exm -\c{c}{\i}}
                    {morning & }
                    {'morning person'}
}

Clearly, this ambiguity cannot be resolved without incorporating into
lexical semantics a Qualia Structure a la Pustejovsky~\cite{puste91}, or
lexical semantic constraints~\cite{fass}. We have been 
incorporating these types of constraints. Unfortunately, descriptive 
work on Turkish linguistics in this regard is very scarce, and there is no
ontology such as Levin's~\cite{levin}.
Using features like [$\mp$animate], [$\mp$artifact],
[$\mp$container], and [$\mp$period], one can
define semantic fields for the derivational morphemes. We expand the set
of features as more lexical items are added to the lexicon. This is a very
labour intensive task; the lack of a large-scale initiative on
lexicography in the manner of LDOCE or COBUILD is hindering the efforts
for automatic extraction of lexical knowledge from on-line resources.

Our strategy
is to obtain complex forms derivationally if the semantic relation
of the bound morpheme to its stem is fairly predictable.
We use lexicalized forms when the meaning is not compositional.
One such case is the denominal verb suffix {\exm -le}, which is
very productive but has no predictable meaning that can be derived
from the lexical semantics of the stem.

\paragraph{Lexical Category Changes:}
As described in section~\ref{sect:intro}, we model the nominal use of 
adjectives in Turkish by a single lexical item which may be
interpreted as a term or a predicate by a lexical rule. There are other
linguistic phenomena that are on the boundary of lexicon and syntax, which
we opted to contain in the lexicon, e.g., non-referential objects, and
valency change in the causatives. In the following, we briefly describe
the lexical rules for them. 

Case assignment is overt in Turkish, which allows for scrambling of the
constituents. All six permutations of the SOV order are felicitous if the
object NP is case marked (e.g., \ref{ex:nonref}a and \ref{ex:nonref}b). 
If the object is non-referential or indefinite (cf.
\ref{ex:nonref}a and \ref{ex:nonref}c), it is not marked
morphologically, which blocks scrambling, and the unmarked SOV order
is used (cf. \ref{ex:nonref}c and \ref{ex:nonref}d). 

\eenumsentence{
\label{ex:nonref}
\item[a.]\shortex{3}{\exm \c{C}ocuk & \exm kitab-{\i} & \exm oku-du}
         { child.NOM & book-ACC(=object) & read-TENSE.3SG}
         {'The child read the book.'}
\item[b.]{\exm Kitab-{\i} \c{c}ocuk  oku-du}
\item[c.]\shortex{3}{\exm \c{C}ocuk & \exm kitap &\exm oku-du}
         {child.NOM & book.ACC & read-TENSE.3SG}
         {'the child read a book ($\cong$ the child did book-reading)'}
\item[d.]{\exm * Kitap  \c{c}ocuk okudu}
}

Non-referential objects are not inflected, and they must occupy the
immediately preverbal position. One way of dealing with nouns, then,
is to keep two entries in the lexicon: one for unmarked form which
may receive case marking and scramble, and one with lexically assigned
case (accusative), which may not scramble. Our solution is
to have a lexical rule that changes the subcategorization
frames of verbs to handle cases where objects may be case-marked NPs
or unmarked Ns. In the second case, the entity is marked indefinite and
all scrambling is blocked by the lexical rule. Figure~\ref{fig:nonrefrule} 
shows the lexical rule in ALE notation (the rule is simplified for
ease of exposition). 

\savebox{\tmp}{
{\scriptsize
\begin{avm}
\[ CAT  \[ HEAD \; 
		\[ {\aS common} \\
		   CASE \; {\aS nominative}
		\]\\
		  ADJUNCTS\|NON-REF \; $+$
	\]\\
   CONT \; \@6
\]
\end{avm}
}
}
\begin{figure}[htb]
%
{\scriptsize
\begin{avm}
\avml
\[ {\aS verb-l} \\
   {\aF SYNSEM} \; \[ {\aF LOCAL\|CAT} \; \[ HEAD \; \@1 \\
					     SUBCAT \; \@2
					  \] \\
		      {\aF NONLOCAL} \;  \@3
		   \]
\] \; $\longmapsto$ \\
\hspace{1cm}
\[ {\aS verb-l} \\
   {\aF SYNSEM} \; \[ {\aF LOCAL\|CAT} \; \[ HEAD \; \@1 \\
					     SUBCAT \; 
			\q< \@4 , \\
			\; \; \usebox{\tmp}
			\q>
					  \]\\
		       NONLOCAL \; \@3
		   \]
\]\; , \\
\hspace{1cm}
{\tt move-object(}\@2,\@4, \[ CAT\|HEAD \; \[ {\aS noun}\\ 
				              CASE \; {\aS accusative}
					   \] \\
			     CONT \; \@6
			  \]{\tt )} 
\avmr
\end{avm}

{\em Where {\tt move-object} is a definite clause which deletes the
accusative object from the {\aF SUBCAT} structure in first argument and
return resulting structure and accusative object in second and third
argument respectively.}  
} 
\caption{Lexical rule for non-referential objects.} 
\label{fig:nonrefrule} 
\end{figure}

Causatives can be modelled in a similar vein. A causative suffix changes
the subcategorization frame of the verb by adding one more argument and
changing the grammatical constraints on the other arguments.
For instance, the new argument becomes the subject (causer), 
and the old subject (agent) is demoted down the grammatical 
hierarchy~\cite{comrie}
to direct object or indirect object, depending on the valency
of the verb:
\eenumsentence{\label{ex:causative}
\item[a.]\shortex{3}{\exm Can & \exm arkada\c{s}-{\i}-n{\i} & \exm \c{c}a\u{g}{\i}r-d{\i}}
           { & friend-POSS-ACC & call-TENSE.3SG}
           {'Can called his friend.'}
\item[b.]\shortex{4}{\exm Mehmet & \exm Can-a & \exm arkada\c{s}-{\i}-n{\i} & 
\exm \c{c}a\u{g}{\i}r-t-t{\i}}
           {  & Can-DAT & friend-POSS-ACC & call-CAUS-TENSE.3SG}
           {'Mehmet had Can call his friend.'}
}

\paragraph{Morphophonemic rules:}
The rules for inflectional and derivational morphology might also
take into account the archiphonemes that are not marked for
certain features. For instance, the locative case marker has
allomorphs {\exm -de/-da/-te/-ta}. They may be represented 
uniquely by two metaphonemes \mbox{-DA} where D is a dental stop unmarked
for voice and A is a low unround vowel unmarked for backness/frontness. 
Vowel harmony and voicing constraints\footnote{cf. 
\cite{kemal:kimmo,murat} for a description
of these processes. \cite{hankamer86} is the original work on Turkish
that combines finite state morphotactics with
morphophonemic alternations.} determine their surface realization during
morphological composition. These kinds of rules are not lexical
rules per se since they do not operate on lexical properties of the
words. In our model, they are embedded in lexical rules for
inflections and derivations.

\section{Conclusion}
For a language with rich morphology,  lexical rules can be used 
for controlled generation of surface forms. Inflections and derivations
can be seen as word-based (local) operations on the root, and thus be
modelled as lexical rules. Phonological alternations in stems
can be embedded in the rules as well. Grammatical role changes, 
type constraints on word subtypes, and noun to NP promotions
(as in non-referential objects) control the proliferation of
lexical entries. 
Semantic contribution of inflections 
seems to be morpheme specific: All derivations take part in 
semantic composition, but some inflections (such as case and causatives)
contribute semantically as well. Most inflections (e.g., person and number
markers), however, have 
grammatical functions only. This is not to say they do not have a  semantic
form, just that in many cases the form is that of identity.
Productive  use of derivations is limited
by the predictability of the semantic relation of the stem to the affix.

We have been testing our lexicon design as part of an HPSG grammar
for Turkish~\cite{onurtez}. The grammar development environment, ALE,
had to be modified to allow run-time evaluation of lexical rules.
Compiling out the lexical rules seems to be impractical, since generating
every possible form for a large lexicon of roots causes exponential
growth in the lexicon. 
Compilation of all surface forms for a lexicon of only 40 root forms
produces around 2800 entries, and takes about 8 minutes on a Sun
Sparcstation 10. 
Run-time execution of rules puts the burden
on parsing or generation. We believe that as the lexicons of NLP systems
become more comprehensive and open-ended, the trade-off will be resolved in 
favour of using the lexical rules on demand at the expense of slower
performance. 

\paragraph{Acknowledgements:}This research is supported in part by grants
from NATO Science Division SfS III (contract TU-LANGUAGE), and Scientific
and Technological Research Council of Turkey (contract EEEAG-90).
\bibliographystyle{plain}
\bibliography{refs}
\end{document}